\documentclass{article}
\usepackage{ijcai20}

\usepackage{times}

\usepackage{soul}
\usepackage{url}
\usepackage[utf8]{inputenc}
\usepackage[small]{caption}
\usepackage{graphicx}
\usepackage{subfigure}
\usepackage{amsmath}
\usepackage{booktabs}
\DeclareMathOperator*{\argmax}{arg\,max}

\usepackage{bm}
\usepackage{arydshln} 
\usepackage{multirow} 
\usepackage{multicol}
\usepackage[dvipsnames]{xcolor}
\usepackage{amssymb}
\usepackage{tikz}
\usepackage{pgfplots}
\usepackage{CJKutf8}
\pgfplotsset{compat=1.14}
\newcommand\citet[1]{\citeauthor{#1} (\citeyear{#1})}
\newenvironment{customlegend}[1][]{%
    \begingroup
    \csname pgfplots@init@cleared@structures\endcsname
    \pgfplotsset{#1}%
}{%
    \csname pgfplots@createlegend\endcsname
    \endgroup
}%
\def\addlegendimage{\csname pgfplots@addlegendimage\endcsname}
\title{Modeling Future Cost for Neural Machine Translation}
\author{
Chaoqun Duan$^1$\thanks{Contribution during internship at NICT.}\and
Kehai Chen$^2$\and
Rui Wang$^2$\and
Masao Utiyama$^2$\and \\
Eiichiro Sumita$^2$\and
Conghui Zhu$^1$\And
Tiejun Zhao$^1$
\affiliations
$^1$Harbin Institute of Technology, Harbin, China \\
$^2$National Institute of Information and Communications Technology (NICT), Kyoto, Japan\\
\emails
cqduan@stu.hit.edu.cn,
\{conghui, tjzhao\}@hit.edu.cn \\
\{khchen, wangrui, mutiyama, eiichiro.sumita\}@nict.go.jp
}
\begin{document}

\begin{CJK*}{UTF8}{gbsn}

\maketitle

\begin{abstract}
Existing neural machine translation (NMT) systems utilize sequence-to-sequence neural networks to generate target translation word by word, and then make the generated word at each time-step and the counterpart in the references as consistent as possible.
However, the trained translation model tends to focus on ensuring the accuracy of the generated target word at the current time-step and does not consider its future cost which means the expected cost of generating the subsequent target translation (i.e., the next target word).
To respond to this issue, we propose a simple and effective method to model the future cost of each target word for NMT systems.
In detail, a time-dependent future cost is estimated based on the current generated target word and its contextual information to boost the training of the NMT model.
Furthermore, the learned future context representation at the current time-step is used to help the generation of the next target word in the decoding.
Experimental results on three widely-used translation datasets, including the WMT14 German-to-English, WMT14 English-to-French, and WMT17 Chinese-to-English, show that the proposed approach achieves significant improvements over strong Transformer-based NMT baseline.
\end{abstract}

\section{Introduction}
\label{Intro}
The future cost estimation plays an important role in traditional phrase-based statistical machine translation (PBSMT)~\cite{koehn2009statistical}.
Typically, it utilizes the pre-learned translation knowledge (i.e., translation model and language model) to compute a future cost of any span of input words in advance for one source sentence.
The computed future cost estimates how hard it is to translate the untranslated part of the source sentence.
For example, for all translation options that have the same number of input words, the higher future cost means that the untranslated part of the source sentence is more difficult to be translated.
During the decoding, PBSMT adds up the partial translation probability score of the current span and its future cost to measure the quality of each translation option.
As a result, a (or several) translation hypothesis, which is extended by translation options with cheaper future cost, is remained in the beam-search stack as the best paths to generate subsequent translation.

Neural machine translation (NMT) systems~\cite{bahdanau2014neural,vaswani2017attention} often utilize sequence-to-sequence neural networks to model translation between the source language and the target language, and achieve state-of-the-art performance on most of the translation tasks~\cite{barrault-etal-2019-findings}.
Compared with the traditional PBSMT, NMT systems model translation knowledge through neural networks.
This means that there is no need to learn large-scale translation rules as traditional PBSMT.
However, lack of translation rules prevents the future cost from being estimated in advance for NMT systems.
Therefore, it is difficult to directly use this effective future cost mechanism in PBSMT to enhance the beam-search stack decoding in NMT systems.

In addition, the NMT systems generally model translation between a source language and a target language in an auto-regressive way, that is, based on the previously translated target word (or context) and the source representation to generate target translation word by word.
However, this makes the trained translation model only focus on ensuring the accuracy of the generated target word at the current time-step and do not consider its future cost as PBSMT.
In other words, there is no mechanism to estimate the future cost of the current generated target word for generating subsequent target translation (i.e., next target word) in NMT systems.

In this paper, we propose a future cost mechanism to learn the expected cost of generating the next target word for NMT systems, for example, state-of-the-art Transformer-based NMT system~\cite{vaswani2017attention}.
Specifically, the future cost is dynamically estimated based on the current target word and its contextual representation instead of pre-estimated in PBSMT.
We then use the estimated future cost to compute an additional loss item to boost the training of the Transformer-based NMT model.
This allows the Transformer-based NMT model to preview the future cost of the current generated target word for the generation of the target word at the next time-step.
In addition, the learned future context representation at the current time-step is further used to help the generation of the next target word in the decoding.
This allows the future cost information to be applied to the beam-search stack decoding in the auto-regressive way instead of in the isolation way in PBSMT, and thereby enhances translation performance of Transformer-based NMT model.

This paper primarily makes the following contributions:
\begin{itemize}
	\item It introduces a novel future cost mechanism to estimate the impact of the current generated target word for generating subsequent target translation (i.e., next target word) in NMT.
	\item The proposed two models can integrate the proposed future cost mechanism into the state-of-the-art Transformer-based NMT system to improve translation performance.
	\item Experiment results on the WMT14 English-to-German, WMT14 English-to-French, and WMT17 Chinese-to-English translation tasks verify the effectiveness and universality of the proposed future cost mechanism.
\end{itemize}

\section{Background}
An advanced Transformer-based NMT model~\cite{vaswani2017attention}, which solely relies on self-attention networks (SANs), generally consists of a SAN-based encoder and a SAN-based decoder. 
Formally, given an source input sequence $\textbf{x}$=\{$x_1$, $\cdots$, $x_J$\} with length of $J$,
this encoder is adopted to encode the source input sequence $\textbf{x}$. 
In particular, each layer includes an SAN sub-layer $\textup{SelfATT}(\cdot)$ and a position-wise fully connected feed-forward network sub-layer $\textup{FFN}(\cdot)$.
A residual connection~\cite{he2016deep} is applied between the SAN sub-layer and the FFN syb-layer, followed by layer normalization $\textup{LN}(\cdot)$~\cite{ba2016layer}.
Thus, the output of the first sub-layer $\textbf{C}_{e}^{n}$ and the second sub-layer $\textbf{H}_{e}^{n}$ are sequentially calculated as Eq.\eqref{eq1:encoder_att} and Eq.\eqref{eq2:encoder_fnn}:
\begin{align}
\label{eq1:encoder_att} \textbf{C}_{e}^{n}&=\textup{LN}(\textup{SelfATT}(\textbf{H}_{e}^{n-1})+\textbf{H}_{e}^{n-1}), \\
\label{eq2:encoder_fnn} 
\textbf{H}_{e}^{n}&=\textup{LN}(\textup{FFN}(\textbf{C}_{e}^{n})+\textbf{C}_{e}^{n}).
\end{align}
Typically, this encoder is composed of a stack of $N$ identical layers.
As a result, $\textbf{H}_{e}^{N}$ is the final source sentence representation to model translation.

Furthermore, this decoder, which is also composed of a stack of $N$ identical layers, models the context information for predicting translations.
In addition to two sub-layers in each decoder layer, the decoder inserts a third sub-layer $\textup{ATT}(\textbf{C}_{i}^{n}, \textbf{H}_{e}^{N})$ perform attention over the output of the encoder $\textbf{H}_{e}^{N}$:
\begin{align}
\label{eq3:decoder_self_att} \textbf{C}_{i}^{n}&=\textup{LN}(\textup{SelfATT}(\textbf{H}_{i}^{n-1})+\textbf{H}_{i}^{n-1}), \\
\label{eq4:decoder_att} \textbf{D}_{i}^{n}&=\textup{LN}(\textup{ATT}(\textbf{C}_{i}^{n}, \textbf{H}_{e}^{N})+\textbf{C}_{i}^{n}), \\
\label{eq5:decoder_fnn} \textbf{H}_{i}^{n}&=\textup{LN}(\textup{FFN}(\textbf{D}_{i}^{n})+\textbf{D}_{i}^{n}).
\end{align}
At the $i$-th time-step, the top layer of the decoder $\textbf{H}_{i}^{N}$ is then used to generate the target word $y_i$ by a linear, potentially multi-layered function (or a softmax function):
\begin{equation}
P(y_i|\textbf{y}_{<i}, \textit{x}) \propto 
\textup{exp}(\textbf{\textit{W}}_\textit{o}\textup{tanh}(\textbf{\textit{W}}_\textit{w}\textbf{H}_{i}^{N}),
\label{eq6:Probabilities_SANs}
\end{equation}
where $\textbf{\textit{W}}_{o}$ and $\textbf{\textit{W}}_{w}$ are projection matrices.
Thus, the cross entropy loss is computed over a bilingual parallel sentence pair $\{[\textbf{x}, \textbf{y}]\}$:
\begin{equation}
\mathcal{L}(\theta)=\argmax_{\theta}\{\sum_{i=1}^{I}\textup{log}P(y_i|\textbf{y}_{<i}, \textbf{x})\}.
\label{eq7:NMT_LossItem}
\end{equation}

\section{Proposed Future Cost Mechanism}
In the traditional PBSMT, the future cost aims to estimate the difficulty of each translation option for one source sentence.
Generally, PBSMT takes the beam-search stack decoding algorithm to select translation options to expand the current hypotheses.
In detail, given a source sentence, all available translation options for any span of input words are collected in advance from the pre-learned translation model and language model.
The future cost of each translation option is then computed based on the statistical scores of the translation model and the language model.
By adding up the partial translation score and the future cost, PBSMT selects a translation option with the cheapest future cost to expand the current translation hypothesis.
Finally, this makes a much better basis for pruning decisions in the beam-search stack decoding.

However, it is difficult to directly apply this future cost of PBSMT to the existing NMT system due to lack of pre-learned translation rules and its auto-regressive characteristic.
Compared with the traditional PBSMT system, the NMT system models the translation knowledge as a time-dependent context vector for translation prediction through large-scale neural networks instead of translation rules.
In particular, the time-dependent context representation is input to Eq.\eqref{eq6:Probabilities_SANs} to compute the translation probability of the target word.
Actually, the translation probability is an important part of computing the future cost in PBSMT.
Meanwhile, NMT is seen as a neural network language model with attention mechanism~\cite{kalchbrenner-blunsom-2013-recurrent,bahdanau2014neural}.
For example, the time-dependent context representation of NMT is regarded as that of the neural network language model~\cite{10.5555/944919.944966}.

Based on the above analysis, we propose a new method to model the future cost for the existing NMT systems. 
Specifically, the proposed approach utilizes the current target word and its context representation to learn a future context representation.
Thus, this future context representation is input to a softmax layer to compute its future cost for the current target word.
Formally, we utilize the top layer $\textbf{H}_{i}^{N}$ learned by the stacked Eq.\eqref{eq1:encoder_att}$\sim$Eq.\eqref{eq5:decoder_fnn} to model the current context information.
The $\textbf{H}_{i}^{N}$ is together with the current generated target word $y_{i}$ to learn a future context representation $\textbf{F}_{i}$ as follows:
\begin{align}
\label{eq8:gru_gate_r} &\textbf{R}_{i}=\sigma(\textbf{W}_{r}\cdot \textbf{E}[y_i]+\textbf{U}_{r}\cdot \textbf{H}_{i}^{N}), \\
\label{eq9:gru_gate_z} &\textbf{Z}_{i}=\sigma(\textbf{W}_{z}\cdot \textbf{E}[y_i]+\textbf{U}_{z}\cdot \textbf{H}_{i}^{N}), \\
\label{eq10:gru_cell} &\textbf{S}_{i}=\textup{ReLU}(\textbf{W}\cdot \textbf{E}[y_i]+\textbf{U}\cdot (\textbf{R}_{i}\odot \textbf{H}_{i}^{N})), \\
\label{eq11:gru_hidden_state} &\textbf{F}_{i}=\textbf{Z}_{i}\odot \textbf{S}_{i}+(1-\textbf{Z}_{i})\odot \textbf{H}_{i}^{N},
\end{align}
where $\textbf{E}$ is the embedding matrix of target vocabulary, $\textbf{W}_{r}$, $\textbf{U}_{r}$, $\textbf{W}_{z}$, $\textbf{U}_{z}$, $\textbf{W}$, and $\textbf{U}$ are model parameters.
$\sigma(\cdot)$ is sigmoid function in which $\odot$ means the element-wise dot.
Note that for the initial future context representation, we use the special source end token ``$<$/s$>$" and the mean of vectors in the source representation $\textbf{H}_{e}^{N}$ as the input to Eq.\eqref{eq8:gru_gate_r}$\sim$Eq.\eqref{eq11:gru_hidden_state} to learn $\textbf{F}_{0}$.

Finally, the learned future context representation $\textbf{F}_{i}$ is as the input to a softmax layer to compute approximate probabilities of temporary target word $\hat{y}_{i+1}$ at the current time-step, called as the future cost of the current generated target word:
\begin{equation}
\hat{P}(\hat{y}_{i+1}|\textbf{y}_{<i}, y_{i}, \textbf{x}) \propto 
\textup{exp}(\bm{\mathcal{W}}_\textit{o}\textup{tanh}(\bm{\mathcal{W}}_\textit{w}\textbf{F}_{i})),
\label{eq12:future_cost} 
\end{equation}
where $\bm{\mathcal{W}}_{o}$ and $\bm{\mathcal{W}}_{w}$ are projection matrices.
Later, $\hat{P}(\hat{y}_{i+1}|\textbf{y}_{<i}, y_{i}, \textbf{x})$ will be used to guide the training of NMT.

\section{Neural Machine Translation with Future Cost Mechanism}
\begin{figure}[th]
    \centering
    \includegraphics[width=3.3in, height=3.0in]{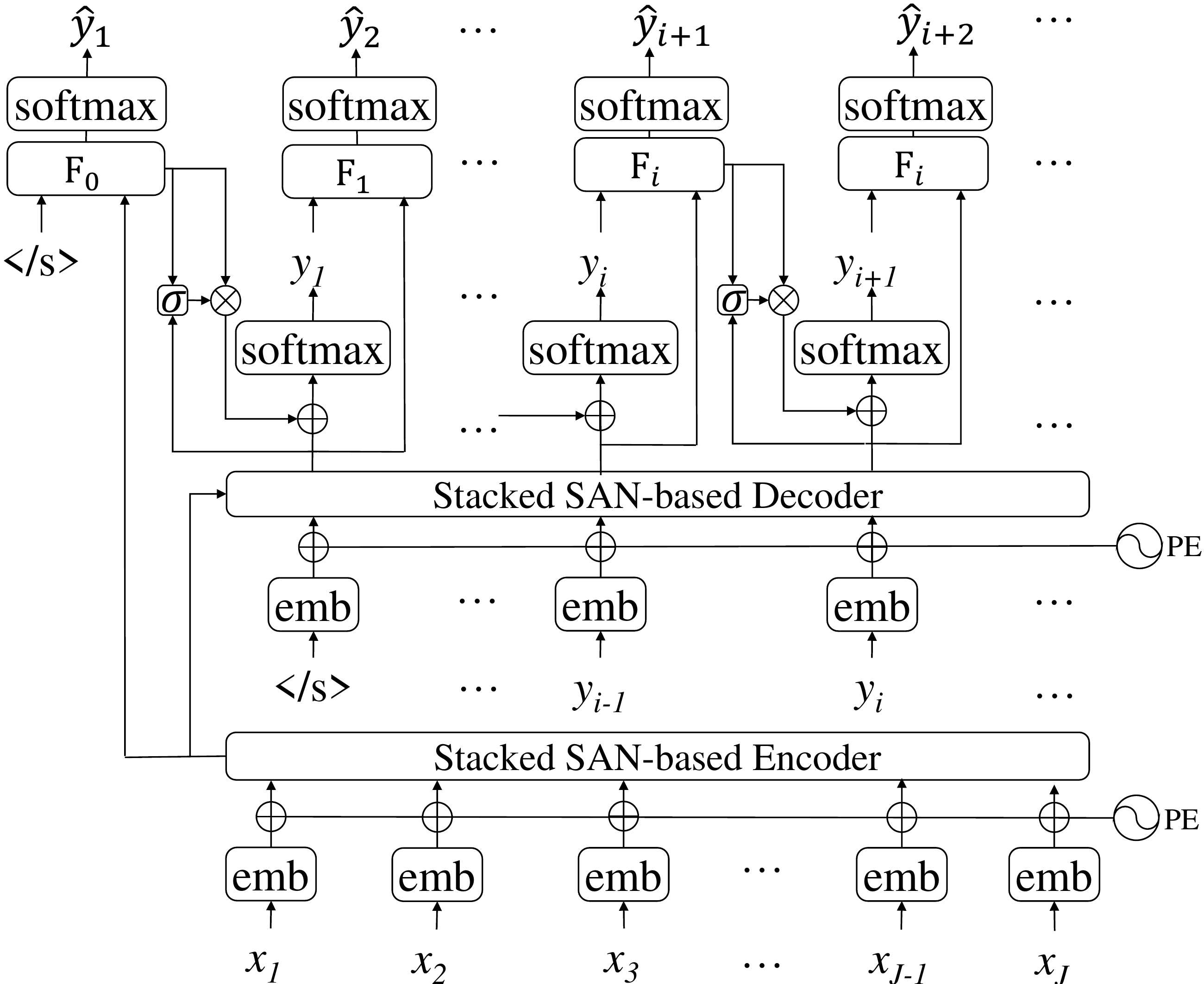} \\
    \caption{The proposed Transformer-based NMT architecture.}
    \label{fig:model}
\end{figure}

In this section, we design two NMT models as Figure~\ref{fig:model} to make use of the proposed future cost in the previous section.
For the first model, we compute an additional loss item of future cost at each time-step, and thereby gain a future cost-ware translation model.
In addition to the additional loss item of future cost, the second model utilizes the learned future context representation to help the generation of target word at the next time-step, thus improving the translation performance of the Transformer-based NMT model.

\subsection{Model I}
The training objective of NMT is to minimize the loss between the words in the translated sentences and those in the references.
Specifically, the word-level cross-entropy between the generated target word by NMT and the reference serves as the loss item at each time-step. 
However, as we analyzed in Section~\ref{Intro}, this existing training objective does not consider the future cost of the current generated target word for generating the next target word.

Therefore, we introduce an addition loss term $\mathcal{F}(\theta)$ to preview the future cost of the generated target word at the current time-step according to Eq.(\ref{eq12:future_cost}):
\begin{equation}
\mathcal{F}(\theta)=\argmax_{\theta}\sum_{i=1}^{I}\textup{log}\hat{P}(\hat{y}_{i+1}|\textbf{y}_{<i}, y_{i}, \textbf{x}; \theta).
\label{eq13:FutureCostAwareLoss}
\end{equation}
The $\mathcal{F}(\theta)$ encourages the translation model to select a target word that is beneficial to the generation of target word at next time-step.
Thus, the loss of the proposed Model I is computed over a bilingual parallel sentence pair $\{[\textbf{x}, \textbf{y}]\}$:
\begin{equation}
\mathcal{J}(\theta)=\mathcal{L}(\theta) + \lambda\cdot\mathcal{F}(\theta),
\label{eq14:FutureCostAwareTraining}
\end{equation}
where $\lambda$ is a hyper-parameter to weight the expected importance of the future cost loss in relation to the trained translation model.
Finally, the trained Model I performs the translation decoding according to Eq.\eqref{eq6:Probabilities_SANs}.

\subsection{Model II}
In PBSMT, the future cost mechanism can help the generation of next target word (or phrase) in addition to minimize search errors in the beam-search stack decoding.
However, the proposed Model I only focuses on learning future cost-aware NMT model to remain optimal translation hypotheses into the search stack.
In other words, this future cost information may be not adequately utilized to predict target translation in NMT.
Therefore, we further make use of the learned future context representation to help the generation of target word at the next time-step.

Formally, at the (i+1)-\textit{th} time-step, the future context representation $\textbf{F}_{i}$ learned at the i-\textit{th} time-step is first concatenated with the top layer of the decoder $\textbf{H}_{i+1}^{N}$ as the input to the sigmoid function to learn a gate scalar $g_{i+1}$:
\begin{equation}
g_{i+1} = \sigma([\textbf{H}_{i+1}^{N}:\textbf{F}_{i}]\textbf{W}_{g}),
\label{eq15:gate}
\end{equation}
where $\sigma$ is a \textup{sigmoid} function and $g_{i+1}$$\in$$[0, 1]$ is used to weight the expected importance of the learned future context representation $\textbf{F}_{i}$ to gain a fused context representation $\overline{\textbf{H}}_{i+1}^{N}$ as follows:
\begin{equation}
\overline{\textbf{H}}_{i+1}^{N} = \textbf{H}_{i+1}^{N}+g_{i+1}\odot \textbf{F}_{i},
\label{eq16:FusedContext}
\end{equation}
where $\textbf{W}_{g}\in \mathbb{R}^{2d_{model}\times 1}$ is a trainable parameter, and $\odot$ is the element-wise dot product.

Finally, the fused context representation $\overline{\textbf{H}}_{i+1}^{N}$ is as the input to a softmax layer to compute translation probabilities of the target word $y_{i+1}$ at the (i+1)-\textit{th} time-step:
\begin{equation}
P(y_{i+1}|\textbf{y}_{<i+1}, \textbf{x}) \propto 
\textup{exp}(\textbf{\textit{W}}_\textit{o}\textup{tanh}(\textbf{\textit{W}}_\textit{w}\overline{\textbf{H}}_{i+1}^{N}).
\label{eq17:Probabilities_Model2} 
\end{equation}
Meanwhile, the training objective of Model II is the same to that of Model I as Eq.(\ref{eq14:FutureCostAwareTraining}).
Note that the future cost is estimated over the ground-truth target word during the training, and is estimated over the generated target word during the decoding.
\begin{table*}[t]
\centering
\begin{tabular}{l|l||l|r|r|l|r|l|r}
\hline \hline
\multicolumn{1}{c|}{\multirow{2}{*}{System}} & \multirow{2}{*}{Architecture} & \multicolumn{3}{c|}{EN-DE}                                & \multicolumn{2}{c|}{EN-FR}                                & \multicolumn{2}{c}{ZH-EN}                                \\ \cline{3-9} 
\multicolumn{1}{c|}{}                        &                               & \multicolumn{1}{c|}{BLEU} & \multicolumn{1}{c|}{\#Speed} & \multicolumn{1}{c|}{\#Param.} & \multicolumn{1}{c|}{BLEU} & \multicolumn{1}{c|}{\#Param.} & \multicolumn{1}{c|}{BLEU} & \multicolumn{1}{c}{\#Param.} \\ \hline \hline
\multicolumn{9}{c}{\textit{Existing NMT systems}}                                                                                                                                                                                                                        \\ \hline
\citet{vaswani2017attention}                  & Trans.base            & 27.30                           & N/A                     & 65.0M                         & 38.10                     & N/A                           & N/A                       & N/A                           \\
\citet{zheng2019dynamic}                  & \;\;\;+Future and Past             & 28.10                           & N/A                     & N/A                        & N/A                     & N/A                           & N/A                       & N/A                           \\
\citet{hao-etal-2019-modeling}                  & \;\;\;+BIARN            & 28.21                           & N/A                     & 97.4M                         & N/A                     & N/A                           & 24.70                       & 117.3M                           \\
\citet{li-etal-2019-information}                  & \;\;\;+Aggregation            & 28.26                           & N/A                     & 92.0M                         & N/A                     & N/A                           & 24.68                       & 112.0M                           \\
\citet{li2020datadependent}                  & \;\;\;+D2GPo            & 27.93                           & N/A                     & N/A                         & 39.23                     & N/A                           & N/A                       & N/A                           \\ \cdashline{1-9}
\citet{vaswani2017attention}                  & Trans.big             & 28.40                           & N/A                     & 213.0M                        & 41.00                     & N/A                           & N/A                       & N/A                           \\
\citet{hao-etal-2019-modeling}                  & \;\;\;+BIARN             & 28.98                           & N/A                     & 333.5M                        & N/A                     & N/A                           & 25.10                       & 373.3M                           \\
\citet{li-etal-2019-information}                  & \;\;\;+Aggregation             & 28.96                           & N/A                     & 297.0M                        & N/A                     & N/A                           & 25.00                       & 337.0M                           \\
\citet{li2020datadependent}                  & \;\;\;+D2GPo             & 29.10                           & N/A                     & N/A                        & 41.77                     & N/A                           & N/A                       & N/A                           \\ \hline \hline
\multicolumn{9}{c}{\textit{Our NMT systems}}                                                                                                                                                                                                                             \\ \hline
\multirow{6}{*}{this work}                    & Trans.base            & 27.42                           & 12.7k                       & 66.5M                           & 39.13                       & 66.9M                           & 23.93                       & 70.7M                           \\
                                              & \;\;\;+Model I        & 27.97+                           & 12.5k                       & 68.4M                           & 39.68+                       & 68.8M                           & 24.48+                       & 72.5M                           \\
                                              & \;\;\;+Model II       & 28.17++                           & 12.0k                       & 68.4M                           & 39.96++                       & 68.8M                           & 24.84++                       & 72.5M                           \\ \cdashline{2-9}
                                              & Trans.big            & 28.45                           & 10.1k                       & 221.1M                           & 41.07                       & 221.9M                           & 24.55                       & 229.4M                           \\
                                              & \;\;\;+Model I                         & 28.98+                           & 9.8k                       & 228.4M                           & 41.79+                       & 229.2M                           & 24.96+                       & 236.8M                           \\
                                              & \;\;\;+Model II                         & 29.12++                           & 9.4k                       & 228.4M                           & 42.02++                       & 229.2M                           & 25.13++                       & 236.8M                           \\ \hline \hline
\end{tabular}
\caption{Results for the WMT14 EN-DE, WMT14 EN-FR, and WMT17 ZH-EN translation tasks. ``\#Speed" denotes the training speed measured in source tokens per second. ``\#Param." indicates the number of model parameters. 
``+/++" after a score indicates that the proposed method was better than the Trans.base model at significance level $p<$0.05/0.01~\protect\cite{collins-koehn-kucerova:2005:ACL}.}
\label{tb1:main_result}
\end{table*}
\section{Experiments}
\subsection{Datasets}
The proposed methods were evaluated on the WMT14 English-to-German (EN-DE), WMT14 English-to-French (EN-FR), and WMT17 Chinese-to-English (ZH-EN) translation tasks.
The EN-DE training set contains 4.5M bilingual sentence pairs, and the \textit{newstest2013} and \textit{newstest2014} data sets were used as the validation and test sets, respectively.
The EN-FR training set contains 36M bilingual sentence pairs, and the \textit{newstest2012} and \textit{newstest2013} datasets were combined for validation and \textit{newstest2014} was used as the test set.
The ZH-EN training set contains 22M bilingual sentence pairs, where the \textit{newsdev2017} and the \textit{newstest2017} data sets were used as the validation and test sets, respectively.
The baselines are involved:

\textbf{Trans.base/big}: a vanilla Transformer-based NMT system without future cost~\cite{vaswani2017attention}, for example Transformer (base) and Transformer (big) models.

\textbf{+Future and Past}~\cite{zheng2019dynamic}: introduce a capsule network into the Transformer NMT system which is adopted to recognize the translated and untranslated contents, and pay more attention to untranslated parts.

Besides, we reported results of the state-of-the-art works~\cite{hao-etal-2019-modeling,li-etal-2019-information,li2020datadependent} for the three translation tasks.

\subsection{Settings}
We implemented the proposed method in the \textit{fairseq}~\cite{ott2019fairseq} toolkit, following most settings in \citet{vaswani2017attention}.
In training the NMT model (base), the byte pair encoding (BPE)~\cite{sennrich-haddow-birch:2016:P16-12} was adopted. The vocabulary size of EN-DE and EN-FR was set to 40K and ZH-EN was set to 32k.
The dimensions of all input and output layers were set to \textit{512}, and that of the inner feedforward neural network layer was set to \textit{2,048}.
The total heads of all multi-head modules were set to \textit{8} in both the encoder and decoder layers. 
In each training batch, there was a set of sentence pairs containing approximately \textit{4,096\(\times\)8} source tokens and \textit{4,096\(\times\)8} target tokens.
The value of label smoothing was set to 0.1, and the attention dropout and residual dropout were $p$ = $0.1$.
We adopt the Adam optimizer \cite{kingma2014adam} to learn the parameters of the model.
The learning rate was varied under a warm-up strategy with warmup steps of 8,000.
For evaluation, we validated the model with an interval of 2,000 batches on the dev set.
Following the training of 300,000 batches, the model with the highest BLEU score for the validation set was selected to evaluate the test sets.
We used multi-bleu.perl\footnote{https://github.com/moses-smt/mosesdecoder/blob/RELEASE-4.0/scripts/generic/multi-bleu.perl} script to obtain the case-sensitive 4-gram BLEU score.
All models were trained on eight V100 GPUs.

\begin{figure*}[ht]
    \centering
    \pgfplotsset{height=5cm,width=6.0cm,compat=1.14,every axis/.append style={thick},legend columns=1}
    \begin{tikzpicture}
        \tikzset{every node}=[font=\small]
        \begin{axis}
        [tick align=inside, legend pos=south east, xticklabels={$(0\text{,}10]$, $(10\text{,}20]$, $(20\text{,}30]$, $(30\text{,}40]$, $(40\text{,}50]$, $(50\text{,})$},
        xtick={0.0,0.2,0.4,0.6,0.8,1.0},
        x tick label style={rotate=45},
        ylabel={BLEU Score},xlabel={Source Sentence Length (EN-DE)}]
        
        \addplot+[sharp plot, mark=triangle*,mark size=1.2pt,mark options={solid,mark color=blue}, color=blue] 
        coordinates
        {(0.0,24.55) (0.2,26.61) (0.4,26.87) (0.6,27.82) (0.8,28.60) (1.0,28.21)};
        \addlegendentry{Trans.base}
        
        \addplot+ [sharp plot, mark=*,mark size=1.2pt,mark options={mark color=green}, color=green]
        coordinates
        {(0.0,27.20) (0.2,27.17) (0.4,27.99) (0.6,27.87) (0.8,29.12) (1.0,28.39)};
        \addlegendentry{\;\;+Model I}
        
        \addplot+ [sharp plot, mark=square*,mark size=1.2pt,mark options={mark color=red}, color=red] 
        coordinates
        {(0.0,27.61) (0.2,27.38) (0.4,27.76) (0.6,28.03) (0.8,29.24) (1.0,28.79)};
        \addlegendentry{\;\;\;+Model II}
        
        \end{axis}
    \end{tikzpicture}
    \begin{tikzpicture}
    \tikzset{every node}=[font=\small]
        \begin{axis}
        [tick align=inside, legend pos=south east, xticklabels={$(0\text{,}10]$, $(10\text{,}20]$, $(20\text{,}30]$, $(30\text{,}40]$, $(40\text{,}50]$, $(50\text{,})$},
        xtick={0.0,0.2,0.4,0.6,0.8,1.0},
        x tick label style={rotate=45},
        ylabel={BLEU Score},xlabel={Source Sentence Length (EN-FR)}]
        
        \addplot+[sharp plot, mark=triangle*,mark size=1.2pt,mark options={solid,mark color=blue}, color=blue] 
        coordinates
        {(0.0,36.98) (0.2,36.78) (0.4,39.17) (0.6,39.76) (0.8,38.97) (1.0,40.25)};
        \addlegendentry{Trans.base}
        
        \addplot+ [sharp plot, mark=*,mark size=1.2pt,mark options={mark color=green}, color=green]
        coordinates
        {(0.0,38.21) (0.2,37.18) (0.4,39.98) (0.6,40.28) (0.8,39.25) (1.0,40.42)};
        \addlegendentry{\;\;+Model I}
        
        \addplot+ [sharp plot, mark=square*,mark size=1.2pt,mark options={mark color=red}, color=red] 
        coordinates
        {(0.0,37.29) (0.2,37.18) (0.4,40.22) (0.6,40.70) (0.8,40.05) (1.0,40.71)};
        \addlegendentry{\;\;\;+Model II}
        
        \end{axis}
    \end{tikzpicture}
    \begin{tikzpicture}
    \tikzset{every node}=[font=\small]
        \begin{axis}
        [tick align=inside, legend pos=south east, xticklabels={$(0\text{,}10]$, $(10\text{,}20]$, $(20\text{,}30]$, $(30\text{,}40]$, $(40\text{,}50])$, $(50\text{,})$},
        xtick={0.0,0.2,0.4,0.6,0.8,1.0},
        x tick label style={rotate=45},
        ylabel={BLEU Score},xlabel={Source Sentence Length (ZH-EN)}]
        
        \addplot+[sharp plot, mark=triangle*,mark size=1.2pt,mark options={solid,mark color=blue}, color=blue] 
        coordinates
        {(0.0,16.29) (0.2,24.49) (0.4,25.20) (0.6,23.71) (0.8,22.43) (1.0,22.24)};
        \addlegendentry{Trans.base}
        
        \addplot+ [sharp plot, mark=*,mark size=1.2pt,mark options={mark color=green}, color=green] 
        coordinates
        {(0.0,16.63) (0.2,24.60) (0.4,26.10) (0.6,24.33) (0.8,23.26) (1.0,23.25)};
        \addlegendentry{\;\;+Model I}
        
        \addplot+ [sharp plot, mark=square*,mark size=1.2pt,mark options={mark color=red}, color=red] 
        coordinates
        {(0.0,17.20) (0.2,24.78) (0.4,26.03) (0.6,25.39) (0.8,23.79) (1.0,23.92)};
        \addlegendentry{\;\;\;+Model II}
        \end{axis}
        
    \end{tikzpicture}
    
    \caption{Trends of BLEU scores with different source length on the EN-DE, EN-FR and ZH-EN test sets}
    \label{fig:TranslationDifferentLength}
\end{figure*}
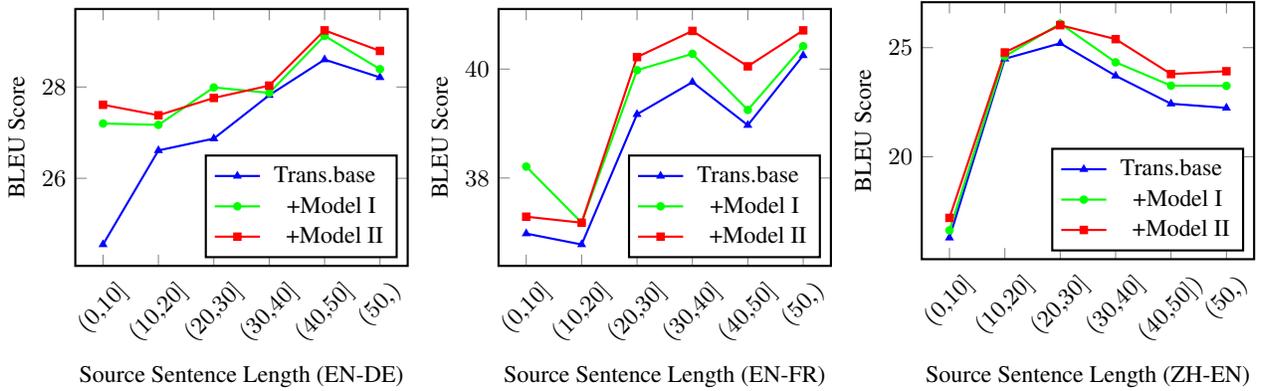
\subsection{Overall Results}
The main results of the translation are shown in Tables~\ref{tb1:main_result}.
We made the following observations:

1) The performance of our implemented Trans.base/big is slightly superior to that of the original Trans.base/big in the EN-DE and EN-FR dataset.
This indicates that it is a strong baseline NMT system and it makes the evaluation convincing.

2) The proposed Model I and Model II significantly outperformed the baseline Trans.base.
This indicated that the future cost information was beneficial for the Transformer-based NMT.
Meanwhile, the Model II outperformed the comparison system +Future and Past~\cite{zheng2019dynamic}, which means that the future cost estimated at the previous time-step is further used to help the generation of the current target word in the decoding.

3) Compared with Model I, Model II achieved a slight advantage on all tasks. 
This means that it is more effective to enhance the translation of the next target word by integrating the learned future hidden representation into the contextual representation of the next word.

4) We also compared our methods with the baseline Trans.big model. 
In particular, the proposed models yielded similar improvements on the three translation tasks, indicating that the proposed future cost mechanism is a universal method for improving
the performance of the Transformer-based NMT model.

5) The proposed models contain approximately $3\%$ additional parameters.
Training and decoding speeds are nearly the same as Trans.base.
This indicates that the proposed method is efficient by only adding a few training and decoding costs.

\subsection{Translating Sentences of Different Lengths}
The proposed future cost mechanism focuses on capturing its future cost for the generated target word at each time-step, thus measuring how good it is to generate the next target word.
Thus, we show the translation performance of source sentences with different sentence lengths, further verifying the effectiveness of our method.
Specifically, we divided each test set into 6 groups according to the length of the source sentence.
Figure~\ref{fig:TranslationDifferentLength} shows the results of the proposed models and Trans.base model on the three translation tasks.
We observed as follows:

1) Model I and Model II were superior to the Trans.base model in almost every length group on all three tasks.
This means that the future cost information capturing by the proposed approach is beneficial to Transformer-based NMT.

2) Compared with Model I, Model II achieved a slight advantage in most groups on each task. 
This indicates that this future cost information also helps the generation of the next target word in addition to that of the current target word. 

3) BLEU scores of all models decreased when the length was greater than 30 over ZH-EN task. 
In contrast, the trend of BLEU scores increased with the sentence length for EN-DE and EN-FR tasks.
We think that NMT may be good at modeling translation between distant language pairs (i.e., ZH-EN) than similar language pairs (i.e., EN-DE and EN-FR). 
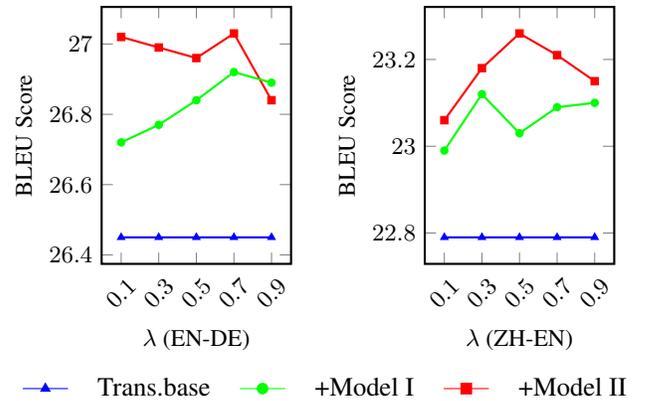
\begin{figure}[h]
    \centering
    \pgfplotsset{height=5cm,width=6.0cm,compat=1.14,every axis/.append style={thick},legend columns=3}
    \begin{tabular}{ccc}
        \begin{tikzpicture}
            \tikzset{every node}=[font=\small]
            \begin{axis}[
            width=4.1cm, enlargelimits=0.13, tick align=inside,
            every axis legend/.append style={at={(0.98,0.50)}},
            every axis legend/.code={\let\addlegendentry\relax},
            xticklabels={$0.1$, $0.3$, $0.5$, $0.7$, $0.9$},
            yticklabel=\pgfmathprintnumber{\tick},
            xtick={0.1,0.3,0.5,0.7,0.9},
            x tick label style={rotate=45},
            ylabel={BLEU Score},xlabel={$\lambda$} (EN-DE)]
            
            \addplot+ [sharp plot, mark=square*,mark size=1.2pt,mark options={mark color=red}, color=red] 
            coordinates
            {(0.1,27.02) (0.3,26.99) (0.5,26.96) (0.7,27.03) (0.9,26.84)};
            \addlegendentry{+Model II}
            
            \addplot+ [sharp plot, mark=*,mark size=1.2pt,mark options={mark color=green}, color=green] 
            coordinates
             {(0.1,26.72) (0.3,26.77) (0.5,26.84) (0.7,26.92) (0.9,26.89)};
            \addlegendentry{+Model I}
        
            \addplot+ [sharp plot, mark=triangle*,mark size=1.2pt,mark options={mark color=blue}, color=blue] 
            coordinates
            {(0.1,26.45) (0.3,26.45) (0.5,26.45) (0.7,26.45) (0.9,26.45)};
            \addlegendentry{Trans.base}
            \end{axis}
        \end{tikzpicture}
        &
        \begin{tikzpicture}
            \tikzset{every node}=[font=\small]
            \begin{axis}[
            width=4.1cm, enlargelimits=0.13, tick align=inside,
            every axis legend/.append style={at={(0.98,0.50)}},
            every axis legend/.code={\let\addlegendentry\relax},
            xticklabels={$0.1$, $0.3$, $0.5$, $0.7$, $0.9$},
            yticklabel=\pgfmathprintnumber{\tick},
            xtick={0.1,0.3,0.5,0.7,0.9},
            x tick label style={rotate=45},
            ylabel={BLEU Score},xlabel={$\lambda$} (ZH-EN)]
            
            \addplot+ [sharp plot, mark=square*,mark size=1.2pt,mark options={mark color=red}, color=red] 
            coordinates
            {(0.1,23.06) (0.3,23.18) (0.5,23.26) (0.7,23.21) (0.9,23.15)};
            \addlegendentry{+Model II}
            
            \addplot+ [sharp plot, mark=*,mark size=1.2pt,mark options={mark color=green}, color=green] 
            coordinates
            {(0.1,22.99) (0.3,23.12) (0.5,23.03) (0.7,23.09) (0.9,23.10)};
            \addlegendentry{+Model I}
    
            \addplot+ [sharp plot, mark=triangle*,mark size=1.2pt,mark options={mark color=blue}, color=blue]
            coordinates
            {(0.1,22.79) (0.3,22.79) (0.5,22.79) (0.7,22.79) (0.9,22.79)};
            \addlegendentry{Trans.base}
            \end{axis}
        \end{tikzpicture} \\
        \multicolumn{3}{c}{
            \begin{tikzpicture}
            \begin{customlegend}[legend columns=3,legend style={align=center,draw=none,column sep=2ex},
                legend entries={\text{Trans.base},
                                \text{+Model I},
                                \text{+Model II},}]
                \addlegendimage{mark=triangle*, blue}
                \addlegendimage{mark=*, green}
                \addlegendimage{mark=square*, red}
            \end{customlegend}
            \end{tikzpicture}
        }
    \end{tabular}
    \caption{BLEU scores of Trans.based, the proposed Model I, and Model II with different $\lambda$ on the EN-DE and ZH-EN validate sets.}
    \label{fig:TranslationDifferentLambda}
\end{figure}
\subsection{Learning Curve of Hyper-parameter $\lambda$}
In Eq.(\ref{eq14:FutureCostAwareTraining}), we introduce a hyper-parameter $\lambda$ to adjust the weight of the future cost loss in relation to the trained translation model.
To tune the value, we conducted experiments with different $\lambda$ on validate set for the three tasks.
As shown in Figure~\ref{fig:TranslationDifferentLambda}, the proposed models achieved an advantage over the Trans.base model with different $\lambda$ on the two tasks.
As a result, for the EN-DE task, Model I and Model II achieved the best BLEU score with $\lambda$=$0.7$ respectively. The trend on EN-FR (not shown) is similar with that on EN-DE and we set $\lambda$=$0.7$ for EN-FR.
For the ZH-EN task, Model I achieved the best BLEU score with $\lambda$=$0.3$ while Model II with $\lambda$=$0.5$.
Finally, the results of Table~\ref{tb1:main_result} are obtained according to these optimized hyper-parameter $\lambda$.

\begin{figure*}[ht]
	\subfigure{
		\begin{minipage}[b]{1.0\linewidth}
            \begin{center}
            \includegraphics[width=5.0in, height=1.0in]{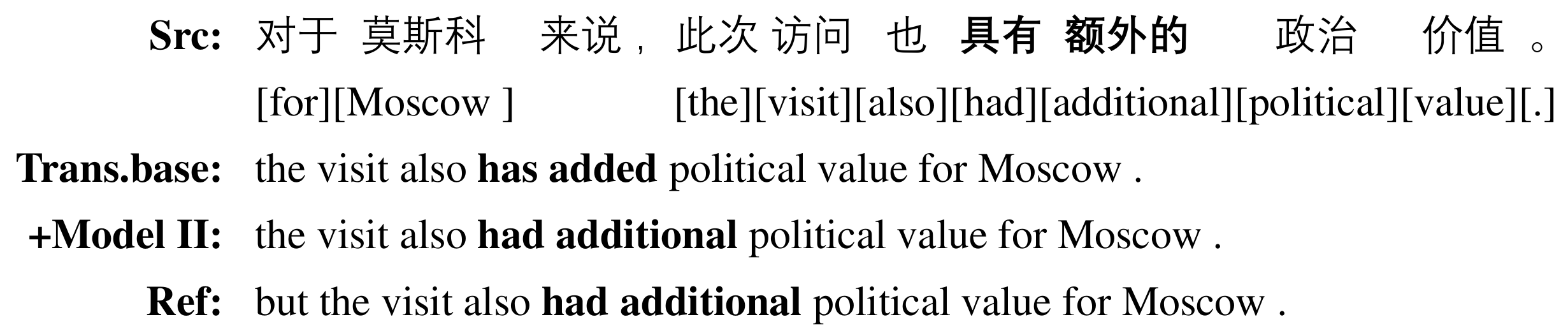}
            \centerline{(a)}
            \end{center}
		\end{minipage}
	}
	\subfigure{
	    \begin{minipage}[b]{1.0\linewidth}
            \includegraphics[width=6.8in, height=0.9in]{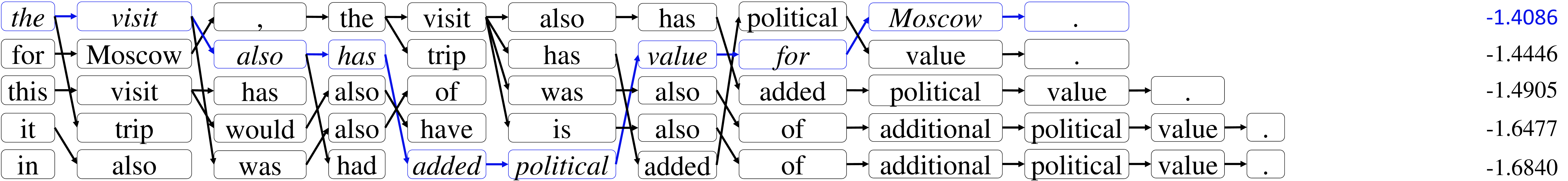} \\
            \centerline{(b)}
		\end{minipage}
	}
	\subfigure{
		\begin{minipage}[b]{1.0\linewidth}
        \includegraphics[width=6.8in, height=0.9in]{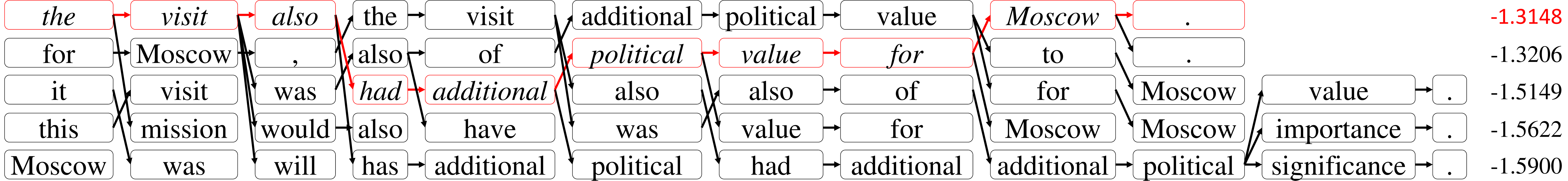} \\
        \centerline{(c)}
        \end{minipage}
	}
	\caption{(a) A translation case; (b) Beam search with beam size=$5$ for Trans.base model; (c) Beam search with beam size=$5$ for +Model II. The end score is translation probability of each decoding path in subfigure (b) and (c), that is, the higher scores denote the better translation.}
    \label{fig:baseline_example}
\end{figure*}
\subsection{Translation Cases}
Figure~\ref{fig:baseline_example}(a) shows a translation case to observe the effect of the proposed method.
Our method translated ``具有[had] 额外的[additional] 政治[political] 价值[value]'' into ``had additional political value'' which are same as the reference, while Trans.base translated it into ``has added political value'' which is different from the reference in the meaning.
We think the reason is that +Model II not only predicts ``had" accurately at the current step, but also captures future cost information that is beneficial for generating ``additional" at the next time-step.
Concretely, Figure~\ref{fig:baseline_example}(b) and Figure~\ref{fig:baseline_example}(c) illustrate the beam-search processing for the Trans.base model and +Model II, respectively. 
In the proposed +Model II, no matter after ``had'' or after ``has", the candidate ``additional'' is produced.
However, in the Trans.base model, only ``added'' follows ``has''. 
This indicates that the learned future contextual representation is beneficial for NMT.

\section{Related Work}
Modeling translated and untranslated information in a source sentence is beneficial to generate target translation in NMT.
\citet{tu-etal-2016-modeling} employed a coverage vector to track the translation part in the source sequence.
Similarly, \citet{mi-etal-2016-coverage} proposed to use a coverage embedding to model the degree of translation for each word in the source sentence.
Later, \citet{li-etal-2018-simple} presented a coverage score to describe to what extent the source words are translated.
Recently, \citet{zheng-etal-2018-modeling} introduced two extra recurrent layers in the decoder to maintain the representations of the past and future translation contents.
Furthermore, \citet{zheng2019dynamic} adopted a capsule network to model the past and future translation contents explicitly.

Compared with the source information, \citet{lin-etal-2018-deconvolution} proposed to adopt a deconvolution network to model the global information of the target sentence.
\citet{NIPS2017_6622} applied a value network to dynamically compute a BLEU score for the rest part of the target sequence based on the difference between the generated sub-sequence and the source sequence. 
\citet{NIPS2017_6775} designed a deliberation network to preview future words through multi-pass decoding.
\citet{li-etal-2018-target} enhanced the attention model by the implicit information of target foresight word oriented to both alignment and translation.
\citet{zhou-etal-2019-synchronous} employed two bidirectional decoders to generate a target sentence in an interactive translation setting. 
Closely related to our work, \citet{weng-etal-2017-neural} proposed to adopt a word prediction mechanism to enhance the contextual representation during the training.
The main difference is that the proposed future cost mechanism can not only minimize search errors but also help the generation of the next word explicitly while \citet{weng-etal-2017-neural} just used it as an extra training objective. 

In short, the above mentioned works focused on adopting the future context to enhance the contextual information of the target word at the current time-step.
Inspired by the future cost in PBSMT, we propose a simple and effective method to estimate the future cost of the current generated target word.
This approach enables the NMT model to preview the translation cost of the subsequent target word at the current time-step and thereby helps the generation of the target word at the next time-step.
This proposed future cost mechanism is integrated into the existing Transformer-based NMT model to improve translation performance.

\section{Conclusion and Future Work}
In this paper, we propose a simple and effective future cost mechanism to enable the translation model to preview the translation cost of next target word at the current time-step. 
We empirically demonstrate that such explicit future cost mechanism benefits NMT with considerable and consistent improvements on three language pairs. 
In the future, we will further extend this work to other NLP tasks.

\bibliography{ijcai20}
\bibliographystyle{named}
\end{CJK*}
\end{document}